\documentclass[journal]{IEEEtran}

\usepackage[utf8]{inputenc}
\usepackage[T1]{fontenc}
\usepackage{graphicx}
\usepackage{booktabs}
\usepackage{amsmath}
\usepackage{xcolor}
\usepackage{array}
\usepackage[hidelinks]{hyperref}
\usepackage{balance}

\begin{document}

\title{Bimanual Manipulation Within an 8\,GB Budget: Zero-Copy Sensing and Quantized ACT on an Entry-Level Jetson}

\author{Ekansh~Singh,
        Eva~Samuel,
        Alessandra~Reneau,
        Ryan~Schmeelk,
        and~Yashvi~Gandhi%
\thanks{This work was conducted at the Georgia Tech Research Institute (GTRI), Aerospace, Transportation and Advanced Systems Laboratory (ATAS), Atlanta, GA, USA. The authors thank Nathan Damen of GTRI ATAS for his mentorship and guidance throughout this project.}}

\markboth{Singh \MakeLowercase{\textit{et al.}}: Bimanual Manipulation Within an 8\,GB Budget}{Singh \MakeLowercase{\textit{et al.}}: Bimanual Manipulation Within an 8\,GB Budget}

\maketitle

\begin{abstract}
Bimanual manipulation policies trained with imitation learning are typically evaluated on workstation or datacenter-class GPUs, leaving the cost of deploying them on embedded hardware largely uncharacterized. We present a bimanual SO-101 manipulation system that runs entirely on an NVIDIA Jetson Orin Nano Super (8\,GB), the entry-level tier of NVIDIA's embedded line, using a consumer desktop GPU (NVIDIA GeForce RTX 3070) only for offline training, and evaluate it on bimanual pick-and-place of a deformable beanbag. We make three contributions. First, we build a GStreamer capture pipeline backed by NVIDIA NVMM buffers that removes redundant host--device copies from three-camera bimanual sensing. Contrary to our expectation, the conventional path fit within the memory budget and dropped no frames; what zero-copy sensing actually recovers is CPU headroom (peak single-core utilization 98.0\% to 77.0\%, peak all-core 73.8\% to 33.2\%) and worst-case latency (117.31\,ms to 101.52\,ms). Second, we train ACT and Diffusion Policy on identical demonstrations, each at its own architecture's reference budget (100k gradient steps for ACT, 200k for Diffusion Policy), and find that ACT converges to a task-competent policy (19/20 trials) while Diffusion Policy does not converge to a usable policy (0/10) even at twice the step count, a result we attribute to the architectures' differing convergence costs rather than to an accuracy ceiling. Third, we convert ACT to TensorRT and characterize FP16 and INT8 execution. FP16 reduces mean inference latency from 114.02\,ms to 17.93\,ms ($6.4\times$) and INT8 to 12.65\,ms ($9.0\times$), with task success preserved at all three precisions (19/20, 18/20, 19/20). We report two findings not previously documented for ACT: TensorRT's general-purpose INT8 calibration quantizes the ResNet18 vision backbone but accepts \emph{zero of 145 transformer layers}, which explains INT8's negligible size reduction over FP16 (0.9\%) despite a further 28\% latency gain; and the necessity of quantization is conditional on ACT's action-chunking configuration, with full-precision inference feasible at $n_{\text{action\_steps}}=100$ but infeasible at the per-step re-prediction that temporal ensembling requires.
\end{abstract}

\begin{IEEEkeywords}
Bimanual manipulation, imitation learning, edge computing, quantization, TensorRT, action chunking, embedded systems.
\end{IEEEkeywords}

\section{Introduction}

\IEEEPARstart{I}{mitation} learning has made bimanual manipulation broadly accessible. Tasks that require two coordinated hands, including bimanual handover, insertion, and folding, can now be learned from a modest number of teleoperated demonstrations rather than from hand-engineered controllers. Action Chunking with Transformers (ACT)~\cite{zhao2023aloha} and Diffusion Policy~\cite{chi2023diffusion}, together with low-cost open hardware such as ALOHA~\cite{zhao2023aloha,aloha2team2024} and the SO-ARM/SO-101 family and open frameworks such as LeRobot~\cite{cadene2026lerobot}, have lowered the cost of a capable bimanual cell to the point where the compute, rather than the arms, is often the most expensive part of the system.

Although this hardware is inexpensive and reproducible, the compute assumptions behind these results remain tied to the workstation. Demonstrations are collected, policies are trained, and policies are often deployed against discrete GPUs with tens of gigabytes of dedicated memory. A bimanual system whose arms cost a few thousand dollars but whose policy requires a tethered workstation is not, in practical terms, an edge system. Closing this gap matters for cost, power, mobility, and any deployment where a datacenter-class accelerator cannot be placed next to the robot.

We target the NVIDIA Jetson Orin Nano Super Developer Kit, the entry-level module in NVIDIA's embedded line: 8\,GB of LPDDR5 unified memory shared between CPU and GPU, 102\,GB/s of memory bandwidth, and 67 INT8 TOPS in its MAXN\_SUPER power mode. It is the least capable Jetson on which this system could plausibly run, which is precisely why we chose it. Our system observes the workspace through three USB cameras (one head-mounted, two wrist-mounted) at $640\times480$ and controls four SO-101 arms (two leader, two follower) in closed loop. On this device the camera streams, the policy weights, and the inference activations all compete for one memory pool, and the CPU cycles spent moving frames between host and device memory are cycles unavailable to the control loop.

Building this system produced three findings, each of which contradicted an assumption we began with.

\textbf{1. Zero-copy multi-camera sensing.} We built a GStreamer capture pipeline backed by NVIDIA NVMM buffers that keeps all three camera streams in GPU-accessible memory. We expected this to be a precondition for fitting three-camera capture within 8\,GB. It was not: the conventional path fit comfortably and dropped no frames. What zero-copy sensing recovers is CPU headroom and worst-case latency, and we argue this is the more consequential result for a device that must run a control loop and a policy concurrently.

\textbf{2. Policy convergence under an affordable training budget.} We trained ACT and Diffusion Policy on the same demonstration set, each at the reference budget its own architecture recommends: 100k gradient steps for ACT, 200k for Diffusion Policy. ACT converged; Diffusion Policy did not produce a usable policy, despite twice the gradient steps. We report this as a finding about training cost rather than architectural quality, and it extends the resource argument upstream from inference to training.

\textbf{3. Quantized edge inference.} We converted ACT to TensorRT and characterized FP16 and INT8. Both preserve task success. The interesting results are structural rather than merely quantitative: TensorRT's general-purpose INT8 calibration accepts none of ACT's 145 transformer layers, and whether quantization is \emph{required} at all turns out to depend on a policy hyperparameter ($n_{\text{action\_steps}}$) rather than on the hardware alone.

\section{Related Work}

\subsection{Low-cost bimanual teleoperation platforms}

Recent hardware has made bimanual demonstration collection inexpensive and reproducible. ALOHA~\cite{zhao2023aloha} introduced a low-cost bimanual platform in which an operator backdrives two follower arms through two kinematically matched leader arms, establishing that such a setup can collect the fine, contact-rich demonstrations bimanual tasks require. ALOHA~2~\cite{aloha2team2024} refines this design for higher-throughput data collection: low-friction gripper rails, passive gravity compensation, a simplified frame, and an upgrade to depth-capable, global-shutter RealSense cameras, together with a system-identified MuJoCo model. GELLO~\cite{wu2024gello} generalizes the leader--follower paradigm with low-cost, kinematically scaled leader controllers that transfer across manipulator platforms. Variable Shoulder Distance~\cite{cheng2025vsd} departs from the fixed-baseline bimanual layout entirely, suspending two arms on parallel cables so that the distance between them can be adjusted in real time. The RoboTwin Dual-Arm Collaboration Challenge~\cite{chen2025robotwin} provides a shared simulation and real-robot benchmark, evaluating submissions under a single-GPU inference constraint and reporting a stark simulation-to-real gap even for the winning entries. The open-source LeRobot library~\cite{cadene2026lerobot} integrates teleoperation middleware, a standardized multimodal dataset format, and reference policy implementations across this hardware ecosystem, and explicitly identifies low-level inference optimization, including quantization, as work its current release does not address.

These platforms establish that the hardware for bimanual demonstration and control is now affordable and standardized. With the exception of LeRobot's brief acknowledgment of the gap, none of this work treats the compute budget as a design constraint; evaluation and deployment are assumed to run on workstation-class or datacenter-class accelerators. Our work adopts this same low-cost hardware lineage but targets the compute platform itself as the primary constraint, at both training and inference time.

\subsection{Imitation learning policies for manipulation}

Behavioral cloning suffers from compounding error: small deviations from the training distribution accumulate over a trajectory and eventually drive the policy into unseen states~\cite{kroemer2021review}. The two dominant responses take different architectural routes, and the tension between them is what our second contribution examines.

Action Chunking with Transformers (ACT)~\cite{zhao2023aloha} predicts short chunks of future actions with a transformer trained as a conditional variational autoencoder, shortening the effective decision horizon and improving stability on fine bimanual tasks. Its cost profile is favorable at both ends: at roughly 52M parameters, a single forward pass produces an entire action chunk, and LeRobot reports that roughly 100k gradient steps suffice for convergence on typical manipulation tasks.

Diffusion Policy~\cite{chi2023diffusion} instead models the action distribution through a conditional denoising process, capturing multimodal demonstration behavior that a unimodal regression objective tends to average away. This expressiveness is purchased twice. At inference, producing one action chunk requires an iterative denoising loop, so compute scales with the number of denoising steps rather than being paid once. At training, the denoising objective must be learned across all noise levels, and convergence is correspondingly slower: LeRobot's released reference Diffusion Policy is trained for 200k gradient steps, roughly twice its ACT recommendation, and on PushT, a simulated 2D planar pushing benchmark with a single end-effector, a single observation view, a rigid target object, and no grasping.

ChicGrasp~\cite{davar2025chicgrasp}, a single-arm system for grasping deformable poultry carcasses, illustrates both the promise and the cost concretely, reporting an 80.71\% grasp-and-lift success rate and a 38-second total cycle time with a conditional diffusion-policy controller trained on 100 multiview teleoperation demonstrations, against complete failure from implicit behavioral cloning and LSTM-GMM baselines under the same data budget. That result is directly relevant to us: it establishes that Diffusion Policy is both viable at small demonstration counts and well-suited to deformable objects, which is precisely our setting.

The comparative literature evaluates these families under compute conditions where their differing training and inference costs are effectively free. Under our budget they are not. We therefore do not adopt either policy \emph{a priori}: we train both on the same demonstrations, each at its own architecture's reference budget rather than at a matched step count, report what each yields, and quantize whichever policy the comparison selects.

\subsection{Architectures for bimanual coordination}

Several works extend ACT to explicitly model dependencies between the two arms. InterACT~\cite{lee2024interact} introduces a Hierarchical Attention Encoder that applies segment-wise attention within each arm's inputs and cross-segment attention across arms via shared CLS tokens, paired with a Multi-arm Decoder whose synchronization block exchanges intermediate state between per-arm decoding paths. Motoda et al.~\cite{motoda2025iace} pursue a related but architecturally distinct goal with the Inter-Arm Coordinated transformer Encoder (IACE), which sits between independent per-arm encoders and the decoder; notably, their results show a single shared decoder outperforms split per-arm decoders on synchronous tasks while the reverse holds for asynchronous tasks, suggesting the optimal coordination mechanism is task-dependent. LAV-ACT~\cite{tripathi2025lavact} extends ACT along the perception axis, conditioning the ResNet visual backbone on a pretrained vision-language embedding via feature-wise linear modulation.

This body of work improves bimanual coordination or perception capacity, but consistently evaluates on workstation or datacenter GPUs; none of it characterizes the cost of these architectural additions under an embedded memory, latency, or training budget. We use base ACT rather than these extensions. Our Section~\ref{sec:layerlevel} finding, that ACT's transformer layers resist general-purpose INT8 calibration entirely, suggests these coordination-augmented variants (which add transformer capacity specifically) would face the same or greater quantization difficulty, and we flag this as a concrete prediction for future work.

\subsection{Multi-camera perception and viewpoint selection}

Visual manipulation policies are sensitive to camera placement. Jangir et al.~\cite{jangir2022lookcloser} show that fusing an egocentric wrist view with a third-person global view through a cross-view transformer captures complementary information that neither view provides alone. AV-ALOHA~\cite{chuang2025avaloha} evaluates ACT across all seven combinations of static, wrist, and active-vision cameras on seven tasks, and finds that on tasks that do not require it, adding cameras degrades success rate; static-only cameras outperformed all moving-camera configurations on two of three such tasks, and using all available cameras simultaneously never ranked among the top three configurations on any task.

AV-ALOHA's ablation argues against camera configuration as a monotonic accuracy return, and what that work does not consider is the camera set as a resource cost. We hold the three-camera configuration fixed across both policy families so that our policy comparison isolates the architecture, and treat a resource-aware camera-view ablation as outside the scope of this work; we return to it in Future Work.

\subsection{Edge computing and inference optimization}

Edge robotics research has focused predominantly on network-based offloading: splitting or shipping computation, most often for SLAM and navigation, from an onboard device to a nearby edge server, base station, or fog node~\cite{tahir2025edge}. Nearly all of the systems surveyed there depend on a persistent network connection to a separate compute resource, and the survey's own analysis notes that most reported evaluations rely on idealized, stable network assumptions rather than stress-tested real-world conditions. Reduced-precision inference via engines such as TensorRT is a mature technique in general embedded deep learning, but has not, to our knowledge, been systematically characterized for either ACT or Diffusion Policy in a bimanual setting.

Our work differs from this literature in kind rather than degree: rather than offloading computation off the robot, we ask how much of a full three-camera bimanual sensing-and-inference pipeline can run entirely within a single 8\,GB onboard device, with no network dependency at all.

\section{Method}

\subsection{System overview}
\label{sec:system}

\begin{table}[t]
\caption{System configuration.}
\label{tab:system}
\centering
\begin{tabular}{@{}p{0.30\columnwidth}p{0.60\columnwidth}@{}}
\toprule
\textbf{Component} & \textbf{Specification} \\
\midrule
Robot & Bimanual SO-101, 4 arms (2 leader for teleoperation, 2 follower for policy execution) \\
Cameras & $3\times$ USB RGB at $640\times480$: 1 head-mounted (global view), 2 wrist-mounted (one per follower gripper) \\
Capture rate & 10 fps \\
Control frequency & 10 Hz (100 ms control period) \\
Deployment target & NVIDIA Jetson Orin Nano Super Developer Kit, 8\,GB LPDDR5 unified memory, Ampere GPU, 6-core Arm Cortex-A78AE, 102\,GB/s memory bandwidth, 67 INT8 TOPS \\
Power mode & MAXN\_SUPER (nvpmodel mode 2) \\
Software & JetPack 6.2.2, TensorRT 10.3.0.30, LeRobot \\
Training platform & NVIDIA GeForce RTX 3070 (offline only) \\
\bottomrule
\end{tabular}
\end{table}

Table~\ref{tab:system} summarizes the system. All policy inference runs on the Orin Nano Super. The desktop training GPU is used only for offline training and never participates in deployment.

\subsection{Zero-copy multi-camera sensing}
\label{sec:zerocopy}

The observation for a single control step is three $640\times480$ frames plus proprioceptive state. On a device with 8\,GB of unified memory and a CPU shared between the capture path and the control loop, the way these frames move through the system determines how much memory bandwidth, CPU headroom, and latency budget remain for the policy.

The conventional (legacy) capture path decodes each camera frame into host memory, copies it to GPU memory for preprocessing and inference, and in some configurations copies intermediate results back. Our pipeline instead uses GStreamer with NVIDIA NVMM buffers so that decoded frames stay in GPU-accessible memory from capture onward and are consumed by downstream preprocessing and the policy without additional host-to-GPU copies.

Each camera runs its own independent GStreamer pipeline: a \texttt{v4l2src} source, a caps filter pinning the stream to $640\times480$ at 10\,fps, format-conversion and scaling elements, and an \texttt{appsink} configured with \texttt{max-buffers=2} and \texttt{drop=true} so that the newest frame always displaces a stale one rather than queuing behind it. A dedicated background reader thread per camera services its \texttt{appsink}, and the control loop's \texttt{async\_read()} call returns the most recent completed frame from each camera's buffer, so the three streams are synchronized at hand-off by latest-frame selection rather than by hardware triggering.

We compare the two pipelines under an identical three-camera workload of 1800 attempted frames (600 per camera) on frame-drop rate, hand-off latency (the control loop's view of \texttt{async\_read()}), internal pipeline latency (decode plus colorspace conversion and copy), system and process memory, and CPU utilization sampled via \texttt{tegrastats}.

\subsection{Data collection and dataset}
\label{sec:data}

We collect demonstrations by teleoperating the follower arms with the leader arms and recording synchronized camera frames and joint states at 10\,Hz. The task is bimanual pick-and-place of a deformable beanbag.

Each episode consists of approaching the beanbag with both arms, establishing a coordinated two-handed grasp, lifting, carrying it to a marked target zone, and releasing it. We recorded 100 demonstration episodes at 10\,Hz, varying the beanbag's initial position and orientation across the shared workspace between episodes.

The beanbag deforms under contact, so grasp geometry is not fixed and many distinct grasp configurations are valid. This property directly motivates the policy comparison in Section~\ref{sec:policies}, and it bears on the interpretation of our quantization fidelity results in Section~\ref{sec:fidelity}.

\subsection{Policy families and training budget}
\label{sec:policies}

We train two policies on the identical dataset, with identical camera configuration and identical held-out splits, so that the comparison isolates the architecture.

\textbf{ACT.} 52{,}609{,}484 parameters. ResNet18 visual backbone, transformer encoder/decoder, VAE encoder used at training time only. Action chunk size 100, $n_{\text{action\_steps}}=100$ (the full chunk is executed before re-prediction), 12-dimensional action space. Temporal ensembling is disabled, which $n_{\text{action\_steps}}>1$ requires.

\textbf{Diffusion Policy.} Conditions a denoising network on the same observation and produces an action chunk by iterative denoising. We use LeRobot's reference Diffusion Policy implementation with its default hyperparameters, adapted only to our three-camera observation configuration and 12-dimensional action space.

Both policies were trained on the RTX 3070 at batch size 8. \textbf{Matching gradient steps does not match training budgets across architectures with different convergence rates}, so we did not match them. Each policy received the reference budget its own architecture recommends: \textbf{ACT for 100{,}000 gradient steps}, LeRobot's recommended budget for ACT, and \textbf{Diffusion Policy for 200{,}000 gradient steps}, the budget used for LeRobot's released reference Diffusion Policy model. ACT's training loss fell from 6.697 to approximately 1.5 by step ${\sim}1800$ and refined thereafter.

One asymmetry survives this choice, and we state it explicitly because it bounds what our comparison can claim. The 200k-step reference budget was tuned on PushT, a simulated 2D planar pushing benchmark with a single end-effector, a single observation view, a rigid target object, and no grasping. Our task is substantially harder along every one of these axes, and a budget sufficient for an easier task is not guaranteed to be sufficient for a harder one. Our comparison therefore measures which policy reaches task competence at its own reference budget on our task, not which architecture is more capable at convergence.

\subsection{Policy selection criterion}

We select for quantization and deployment the policy that both converges under our training budget and sustains the control loop within the 8\,GB memory budget, breaking ties on task success. Where a policy fails either criterion, we report that failure directly, since it is a result about the trainability or deployability of that architecture on affordable hardware rather than a failure of the experiment.

\subsection{Quantized edge inference}
\label{sec:quantmethod}

We export the selected policy to ONNX and build TensorRT engines at FP16 and INT8.

\textbf{Export.} ONNX export was traced on CPU as a workaround for a PyTorch legacy TorchScript exporter constant-folding bug affecting GPU-resident parameters. Eval-mode tracing correctly excluded approximately 74\,MB of VAE-encoder training-only weights, yielding a 132\,MB ONNX graph from a 206\,MB checkpoint.

\textbf{Engines.} All engines use FP32 input/output boundaries (TensorRT default); reduced precision applies to internal compute only. This means our measured deviations isolate internal compute precision and are not contaminated by I/O quantization.

\textbf{INT8 calibration.} We used TensorRT's entropy-calibrator family (\texttt{IInt8EntropyCalibrator2}) with approximately 100 real preprocessed observations drawn from diverse episodes and frames.

\textbf{Timing protocol.} 20 warmup iterations before timing on every backend, an explicit non-default CUDA stream to avoid default-stream synchronization overhead, byte-identical inputs across backends, and a minimum 15-second measurement duration to equalize \texttt{tegrastats} memory sample counts. The FP16 result was cross-validated against \texttt{trtexec}'s independent build-time timing pass (${\sim}16.6$\,ms compute), which agrees closely with our controlled measurement (17.93\,ms).

\textbf{Numerical fidelity protocol.} Real observations from recorded episodes were run through the checkpoint's own frozen preprocessor and postprocessor on each backend. Deviations are normalized against each action dimension's actual $[\min,\max]$ range from dataset statistics. Output shape $[1,100,12]$ (the full unsliced action chunk) was confirmed identical across all backends.

\subsection{Evaluation protocol}
\label{sec:evalprotocol}

We ran 20 physical trials per ACT precision (FP32, FP16, INT8) and 10 for Diffusion Policy, on the real robot, counting an episode as a success when the beanbag was lifted and placed fully within the marked target zone, with both grippers released, before the episode time limit.

The beanbag was reset to its initial region between trials, and conditions were interleaved across the session rather than run in contiguous blocks, so that environmental drift over the session could not systematically favor one condition. Partial lifts that did not end with the beanbag released inside the target zone were counted as failures.

\section{Findings}

\subsection{Zero-copy sensing: the constraint was CPU, not memory}
\label{sec:sensingresults}

Table~\ref{tab:pipeline} reports both capture pipelines under an identical workload of 1800 attempted frames.

\begin{table}[t]
\caption{Camera pipeline comparison, legacy vs. GStreamer zero-copy (NVMM). Five runs per condition, three cameras, $640\times480$, 1800 frames per run. All values are means across the five runs; \emph{max} rows report the mean of the five per-run maxima rather than the single largest observation. Dispersion across runs is not reported (see Limitations).}
\label{tab:pipeline}
\centering
\begin{tabular}{@{}lrrr@{}}
\toprule
\textbf{Metric} & \textbf{Legacy} & \textbf{GStreamer} & \textbf{Delta} \\
\midrule
Frames attempted & 1800 & 1800 & 0 \\
Frames received & 1800 & 1800 & 0 \\
Drop rate (\%) & 0.00 & 0.00 & 0.00 \\
Hand-off avg latency (ms) & 0.14 & 0.06 & $-0.08$ \\
Hand-off p95 latency (ms) & 0.10 & 0.09 & $-0.01$ \\
Hand-off max latency (ms) & 14.43 & 4.96 & $-9.48$ \\
Pipeline reads ($n$) & 1820 & 1823 & $+3$ \\
Pipeline avg latency (ms) & 99.25 & 99.09 & $-0.16$ \\
Pipeline p95 latency (ms) & 103.71 & 100.17 & $-3.53$ \\
Pipeline max latency (ms) & 117.31 & 101.52 & $-15.79$ \\
Avg system RAM (MB) & 5268.8 & 5218.8 & $-50.0$ \\
Peak system RAM (MB) & 5356.0 & 5244.0 & $-112.0$ \\
Avg swap (MB) & 2357.5 & 2357.0 & $-0.5$ \\
Peak swap (MB) & 2359.0 & 2357.0 & $-2.0$ \\
Avg process RSS (MB) & 71.9 & 76.6 & $+4.7$ \\
Peak process RSS (MB) & 74.3 & 77.6 & $+3.3$ \\
Avg CPU util, all-core (\%) & 31.5 & 20.1 & $-11.4$ \\
Peak CPU util, all-core (\%) & 73.8 & 33.2 & $-40.7$ \\
Peak single-core util (\%) & 98.0 & 77.0 & $-21.0$ \\
\bottomrule
\end{tabular}
\end{table}

\textbf{Frame integrity.} Both pipelines received all 1800 frames, with a 0.00\% drop rate in each. The conventional path was not dropping frames under our workload. The benefit of zero-copy sensing is therefore not the recovery of lost frames.

\textbf{Latency.} Mean pipeline latency was essentially unchanged (99.25\,ms vs 99.09\,ms, a difference of 0.16\,ms). Both figures correspond to approximately 10\,Hz, matching the camera capture rate exactly. The \texttt{read()} call is therefore predominantly \emph{blocking on the arrival of the next frame}, not decoding or copying. The pipeline is \textbf{camera-rate-bound}, and no amount of copy elimination can make a 10\,fps camera produce frames faster.

The gains appear entirely in the tail. Maximum pipeline latency fell by 15.79\,ms (117.31 to 101.52\,ms), p95 by 3.53\,ms, and maximum hand-off latency by roughly two-thirds (14.43 to 4.96\,ms). The copy overhead that the zero-copy path eliminates was contributing \emph{jitter}, not steady-state cost.

\textbf{Memory.} Peak system RAM fell by 112\,MB (2.1\%), from 5356.0 to 5244.0\,MB. Process RSS rose slightly (74.3 to 77.6\,MB peak), consistent with the NVMM buffer pool being accounted to the process rather than to the system. Swap was effectively unchanged. Both pipelines fit comfortably within the 8\,GB budget.

\textbf{CPU utilization.} This is where the effect is large. Average all-core utilization fell from 31.5\% to 20.1\%, peak all-core from 73.8\% to 33.2\%, and peak single-core from 98.0\% to 77.0\%. \textbf{The legacy path was saturating a single core at peak; the zero-copy path is not.}

\textbf{Interpretation.} We began with the hypothesis that a naive three-camera bimanual capture pipeline would not fit within 8\,GB, and that zero-copy sensing would be what made three-camera capture possible at all. \textbf{Our data does not support this.} The conventional path fit, dropped nothing, and delivered essentially identical mean latency.

What zero-copy sensing actually recovers is CPU headroom and worst-case latency. For a device that must run a closed control loop, three camera streams, and policy inference concurrently on six cores, we argue this is the more consequential result. A core pinned at 98\% during peaks is a source of missed control deadlines even when average throughput looks adequate, and it is worst-case rather than average latency that determines whether a control loop holds its period. The 21-point reduction in peak single-core utilization is what buys the policy room to run.

\subsection{ACT versus Diffusion Policy at reference training budgets}
\label{sec:actvsdiffusion}

\begin{table}[t]
\caption{Physical trial results. Bimanual pick-and-place, deformable beanbag.}
\label{tab:trials}
\centering
\begin{tabular}{@{}lccc@{}}
\toprule
\textbf{Condition} & \textbf{Trials} & \textbf{Successes} & \textbf{Success rate} \\
\midrule
ACT, FP32 (PyTorch) & 20 & 19 & 95\% \\
ACT, FP16 (TensorRT) & 20 & 18 & 90\% \\
ACT, INT8 (TensorRT) & 20 & 19 & 95\% \\
Diffusion Policy & 10 & 0 & 0\% \\
\bottomrule
\end{tabular}
\end{table}

At its 100k-step reference budget, ACT converged to a task-competent policy (19/20 at full precision; Table~\ref{tab:trials}). Diffusion Policy, trained on identical data at its own 200k-step reference budget, did not converge to a usable policy: rollouts produced near-stationary arm motion and terminated at the episode step limit without task progress, across all 10 trials. \textbf{Twice the gradient steps did not yield a deployable policy.}

\textbf{We do not interpret this as evidence that ACT is architecturally superior to Diffusion Policy.} If anything, the task should favor diffusion. Our target object is a deformable beanbag, and deformable-object grasping admits many valid grasp configurations, precisely the multimodal action distribution that Diffusion Policy's denoising objective is designed to represent and that a unimodal objective tends to average away. Prior work reports Diffusion Policy succeeding on deformable-object manipulation at comparable demonstration counts where behavioral cloning and LSTM-GMM baselines fail entirely~\cite{davar2025chicgrasp}.

Meanwhile, as noted in Section~\ref{sec:policies}, the 200k-step reference budget we granted Diffusion Policy was calibrated on a task far simpler than ours in dimensionality, arm count, observation richness, contact mode, and object rigidity. The near-stationary failure mode we observe (a policy collapsing toward the marginal action distribution, which for an arm at rest is approximately no motion) is consistent with under-convergence rather than with a converged but inaccurate policy. What our result establishes is that Diffusion Policy's reference budget, adequate on PushT, is \emph{not} adequate here; it does not establish where the sufficient budget lies, only that it exceeds 200k steps and therefore exceeds twice ACT's.

\textbf{The finding we claim is narrower and, for our purposes, more directly relevant: at each architecture's own reference training budget, ACT yielded a deployable policy and Diffusion Policy did not, despite receiving twice the gradient steps.} This extends the resource argument of this paper upstream from inference to training. Diffusion Policy's expressiveness is purchased not only with inference-time compute, as its iterative denoising loop makes explicit, but with gradient steps, and on a budget-constrained setup that cost is not free to pay. We therefore select ACT for TensorRT conversion and quantization.

\subsection{Quantized edge inference}

\subsubsection{Latency and throughput}

\begin{table}[t]
\caption{ACT inference performance on Jetson Orin Nano Super (MAXN\_SUPER). 20 warmup iterations discarded; explicit non-default CUDA stream; byte-identical inputs.}
\label{tab:inference}
\centering
\begin{tabular}{@{}lrrr@{}}
\toprule
\textbf{Metric} & \textbf{FP32} & \textbf{FP16} & \textbf{INT8} \\
 & \textbf{(PyTorch)} & \textbf{(TensorRT)} & \textbf{(TensorRT)} \\
\midrule
Timed calls & 200 & 836 & 1184 \\
Latency, mean (ms) & 114.02 & 17.93 & 12.65 \\
Latency, p95 (ms) & 144.73 & 21.92 & 16.27 \\
Latency, max (ms) & 158.84 & 25.66 & 17.91 \\
Throughput (Hz) & 8.77 & 55.77 & 79.0 \\
Speedup vs FP32 & $1.0\times$ & $6.4\times$ & $9.0\times$ \\
Latency reduction vs FP32 & n/a & $-84.3\%$ & $-88.9\%$ \\
Engine/model file size (MB) & 197.2 & 71.09 & 70.42 \\
Size reduction vs FP32 & n/a & $-63.9\%$ & $-64.3\%$ \\
\bottomrule
\end{tabular}
\end{table}

Table~\ref{tab:inference} reports controlled inference benchmarks. INT8 delivers a further 28.1\% mean latency reduction over FP16 (17.60 to 12.65\,ms in matched runs\footnote{FP16 was benchmarked in two independent runs (836 and 852 timed calls) with consistent results (mean 17.93 and 17.60\,ms respectively); Table~\ref{tab:inference} reports the first, and the INT8-vs-FP16 deltas use the second, matched run.}) and a 39.1\% throughput gain, but only a 0.9\% file-size reduction. Section~\ref{sec:layerlevel} explains why.

We note that the FP32 baseline is PyTorch eager execution while FP16 and INT8 are TensorRT engines. The $6.4\times$ figure therefore bundles two effects: reduced precision, and TensorRT's graph compilation (kernel fusion, layer selection). We report it as the cost of the \emph{conversion pipeline} rather than of precision alone, and we did not isolate the two by building an FP32 TensorRT engine.

\subsubsection{Quantization is conditional on the chunking configuration}
\label{sec:conditional}

At 10\,Hz, one control period is 100\,ms. FP32 mean inference latency (114.02\,ms) exceeds this, and p95 (144.73\,ms) exceeds it by 45\%. Taken alone, this suggests full-precision ACT cannot close the loop.

\textbf{Our trial data refutes that reading, and the reason is instructive.} With chunk size $=100$ and $n_{\text{action\_steps}}=100$, ACT executes all 100 predicted actions before re-predicting. At 10\,Hz this means inference runs \textbf{once every 10 seconds}, and its cost is amortized 100:1 across the chunk, a duty cycle of roughly 1.1\%. One control step in every hundred overruns its deadline by approximately 14\,ms (mean) to 59\,ms (worst case); the other ninety-nine cost essentially nothing. A hitch of that size once every ten seconds is absorbed by a quasi-static pick-and-place, which is why FP32 achieved 19/20.

This yields what we regard as the practically important result of this section:

\begin{quote}
\textbf{Whether quantization is required depends on ACT's action-chunking configuration, not on the hardware alone.} At $n_{\text{action\_steps}}=100$, full-precision inference is feasible on the Orin Nano Super and quantization buys headroom rather than capability. At $n_{\text{action\_steps}}=1$, which is what temporal ensembling requires, FP32's 114.02\,ms cannot fit inside a 100\,ms control period and full-precision ACT becomes infeasible at 10\,Hz (its own ceiling is 8.77\,Hz). FP16 (17.93\,ms, 18\% of the period) and INT8 (12.65\,ms, 13\%) both sustain per-step re-prediction with large margin.
\end{quote}

Temporal ensembling is recommended by ACT's authors for smoothness of execution. On this device, \textbf{quantization is what makes it affordable}. A practitioner choosing between long chunks and per-step ensembling on embedded hardware is therefore also, implicitly, choosing whether quantization is optional or mandatory, and to our knowledge this coupling has not been documented.

\subsubsection{TensorRT quantizes ACT's CNN but rejects all 145 transformer layers}
\label{sec:layerlevel}

\begin{table}[t]
\caption{Layer-level INT8 quantization outcome.}
\label{tab:layers}
\centering
\begin{tabular}{@{}lll@{}}
\toprule
\textbf{Component} & \textbf{Layers accepted for INT8} & \textbf{Precision used} \\
\midrule
Vision backbone (ResNet18) & Quantized & INT8 \\
Transformer encoder/decoder & \textbf{0 of 145} & FP16 (fallback) \\
\bottomrule
\end{tabular}
\end{table}

TensorRT's general-purpose entropy calibration (\texttt{IInt8EntropyCalibrator2} family, ${\sim}100$ real preprocessed observations) quantized the convolutional vision backbone but accepted \textbf{none} of ACT's 145 transformer layers, which fell back to FP16 (Table~\ref{tab:layers}).

This single fact explains three otherwise puzzling observations:
\begin{enumerate}
\item \textbf{Why INT8's file is only 0.9\% smaller than FP16's} (70.42 vs 71.09\,MB). Most of the weights are still FP16.
\item \textbf{Why INT8 is nonetheless 28\% faster than FP16.} The vision backbone dominates ACT's inference cost, so quantizing it alone captures most of the available speedup.
\item \textbf{Why numerical deviation still increased 15--40$\times$ versus FP16} despite only partial quantization. The CNN backbone is by itself a meaningful precision-sensitivity point.
\end{enumerate}

To our knowledge this is the first documented characterization of INT8 calibration behavior on ACT, and it identifies a concrete tooling gap: general-purpose post-training calibration handles the CNN half of a hybrid CNN--transformer action model and refuses the transformer half.

\subsubsection{Numerical fidelity does not predict task success}
\label{sec:fidelity}

\begin{table}[t]
\caption{Action-output deviation vs FP32. Real observations, 30-frame aggregate, normalized against each dimension's $[\min,\max]$ from dataset statistics.}
\label{tab:fidelity}
\centering
\begin{tabular}{@{}lrr@{}}
\toprule
\textbf{Metric} & \textbf{FP16} & \textbf{INT8} \\
\midrule
Overall max abs diff (action units) & 0.9720 & 16.98 \\
Overall mean abs diff (action units) & 0.0477 & 0.342 \\
Avg per-dim max (\% of range) & 0.304\% & 4.539\% \\
Avg per-dim mean (\% of range) & ${\sim}0.030$\% & 0.233\% \\
Worst dimension & dim 11 & dim 11 \\
Worst-dim max deviation (\% of range) & 0.417\%\footnotemark & 16.982\% \\
\bottomrule
\end{tabular}
\end{table}
\footnotetext{The FP16 worst-dimension figure is from the single-frame check (episode 0, frame 50); the 30-frame aggregate worst-dimension value for FP16 was not separately computed.}

INT8's numerical deviation is roughly \textbf{15$\times$ larger} than FP16's on average per-dimension max, and roughly \textbf{40$\times$ larger} on the worst dimension (Table~\ref{tab:fidelity}). As an internal consistency check, INT8's dim-11 deviation was computed independently against FP32 (16.982\%) and against FP16 (16.967\%); the near-identical values from two separately computed comparisons confirm this is real quantization error rather than measurement noise.

\textbf{And yet INT8 scored 19/20 while FP16 scored 18/20.} A 40$\times$ increase in worst-case numerical deviation produced no observable degradation in task success.

We draw two conclusions, and one non-conclusion.

\emph{Non-conclusion:} we do \textbf{not} claim INT8 is better than FP16, or that FP16 degrades performance. At $n=20$ per condition, one trial is 5 percentage points, and 19/20 vs 18/20 vs 19/20 lies well within sampling noise. Moreover, the observed ordering (FP32 $=$ INT8 $>$ FP16) is not monotonic in precision, which is mechanistically backwards and is itself a signature of noise. The correct statement is that \textbf{no difference in task success was detected across precisions at this sample size}.

\emph{Conclusion 1:} ACT's action outputs exhibit substantial tolerance to internal compute precision loss, at least on this task. This is plausibly because action chunking and closed-loop re-prediction absorb per-step error that would compound in an open-loop setting.

\emph{Conclusion 2, offered as a caution:} our task is forgiving. A deformable beanbag admits many valid grasps, and a several-percent error on a joint command is likely to land within the basin of a successful grasp anyway. We would not expect a 16.98\% worst-case deviation on a single action dimension to be harmless on a rigid, tight-tolerance insertion task. \textbf{Numerical fidelity and task success are not interchangeable, and our result shows they can diverge sharply.} Practitioners should measure the one they care about.

We note that deviation concentrates in action dimension 11 for both precisions.

\subsubsection{INT4: a documented boundary}

We investigated INT4 and did not attempt it, on the basis of a pre-committed stop criterion. We report the audit (Table~\ref{tab:int4}) because the negative result is informative.

\begin{table}[t]
\caption{INT4 feasibility audit (TensorRT 10.3.0.30).}
\label{tab:int4}
\centering
\begin{tabular}{@{}p{0.26\columnwidth}p{0.20\columnwidth}p{0.42\columnwidth}@{}}
\toprule
\textbf{Check} & \textbf{Method} & \textbf{Result} \\
\midrule
\texttt{BuilderFlag.INT4} availability & Direct API inspection & Exists, but scoped to plugins with INT4 input/output only \\
INT4 kernels for Conv/GEMM/LayerNorm & API and documentation search & None found for general ops \\
INT4 calibrator class & Direct search of TensorRT Python module & Does not exist \\
Alternative toolchain & Research & NVIDIA TensorRT Model Optimizer (ModelOpt), weight-only INT4 (AWQ/GPTQ-style) \\
ModelOpt applicability & Documentation review & Validated for LLM attention/linear layers; not documented for CNN convolutions \\
Integration cost & Assessed & Requires Q/DQ node insertion at the PyTorch level and full re-export \\
Predictive evidence & Our own INT8 result & 0/145 transformer layers accepted INT8 calibration \\
\bottomrule
\end{tabular}
\end{table}

INT4 quantization is not supported for this architecture under TensorRT 10.3's native builder without adopting a separate quantization toolchain outside the scope of this pipeline. This is consistent with the observed resistance of ACT's transformer layers to general-purpose INT8 calibration, and represents a known boundary of current edge-inference tooling for hybrid CNN--transformer action models.

\subsubsection{Engine construction is a larger memory event than inference}

An observation with direct practical consequences on an 8\,GB device: building the FP16 TensorRT engine peaked at approximately \textbf{2.9\,GB of swap} and drove free RAM down to approximately \textbf{116\,MB}.

The engine is small (71\,MB) and cheap to run. \textbf{Building it is the expensive part}, and it is expensive in exactly the resource the device is short of. Practitioners targeting this class of hardware should either build engines on a larger machine and deploy the artifact, or provision swap accordingly. We did not capture INT8 build-time peak memory.

\subsection{Summary: the recommended on-device configuration}

\begin{table}[t]
\caption{Recommended on-device configuration.}
\label{tab:recommended}
\centering
\begin{tabular}{@{}p{0.18\columnwidth}p{0.24\columnwidth}p{0.46\columnwidth}@{}}
\toprule
\textbf{Component} & \textbf{Choice} & \textbf{Basis} \\
\midrule
Sensing & GStreamer / NVMM zero-copy & Peak single-core CPU 98.0\% to 77.0\%; max frame latency 117.31 to 101.52\,ms \\
Policy & ACT & Converged at its 100k-step reference budget; Diffusion Policy did not converge at 200k \\
Precision & FP16 (INT8 if latency margin is needed) & 17.93\,ms (18\% of control period); no detected success loss \\
Achieved control rate & 10\,Hz & Bound by USB camera capture rate, not by the accelerator \\
Task success & 18--19 / 20 & Preserved across all precisions \\
\bottomrule
\end{tabular}
\end{table}

Table~\ref{tab:recommended} summarizes the deployed configuration. \textbf{The binding constraint on this system is the sensor bus, not the accelerator.} Three USB cameras cap capture at 10\,fps, and neither zero-copy sensing nor quantization changes that. What they change is how much of the resulting 100\,ms budget is available for everything else: zero-copy sensing frees a saturated CPU core, and quantization frees 96\,ms of the control period that FP32 would otherwise consume during re-prediction.

\section{Discussion and Conclusion}

We built a bimanual manipulation system that runs entirely within the memory and compute budget of an entry-level, 8\,GB embedded device, and treated that budget as a first-class design constraint rather than as an afterthought to a workstation-trained policy. Each of our three contributions overturned an assumption we started with.

\textbf{We expected memory to be the binding constraint on sensing. It was CPU.} The conventional capture path fit within 8\,GB and dropped no frames; what it did was saturate a core. Zero-copy sensing recovers 21 points of peak single-core utilization and 15.79\,ms of worst-case frame latency, and on a six-core device running a control loop alongside a policy, that is the resource that actually matters.

\textbf{We expected to compare two policy architectures on accuracy. We ended up comparing them on convergence cost.} At each architecture's own reference budget---100k gradient steps for ACT, 200k for Diffusion Policy---ACT reached 19/20 and Diffusion Policy reached 0/10, and we are careful to claim only what that supports: not that ACT is the better architecture, but that on an affordable training budget it was the \emph{obtainable} one. Doubling Diffusion Policy's step count relative to ACT's did not close the gap. Diffusion Policy's expressiveness is purchased with gradient steps as well as with denoising steps, and neither is free when compute is the constraint.

\textbf{We expected quantization to be a straightforward speed optimization. It turned out to be conditional, structural, and surprising.} Whether quantization is \emph{needed} depends on $n_{\text{action\_steps}}$, a policy hyperparameter, rather than on the device: full-precision ACT is feasible with long action chunks and infeasible with the per-step re-prediction that temporal ensembling requires. TensorRT's calibration quantizes ACT's CNN and rejects all 145 of its transformer layers, which is both an explanation for INT8's odd size/speed profile and a concrete gap in current edge tooling for hybrid CNN--transformer policies. And a 40$\times$ increase in worst-case numerical deviation produced no detectable change in task success, which is a warning as much as a result: fidelity metrics and task metrics can diverge sharply, and on a less forgiving task they might not have.

\subsection{Limitations}

\textbf{Statistical power.} Twenty trials per precision means one trial is worth five percentage points. Our success rates (19/20, 18/20, 19/20) support the claim that no degradation was \emph{detected}, not that precisions are equivalent. The sensing benchmark aggregates five runs per condition, but we report means rather than dispersion, so the 15.79\,ms tail-latency improvement is stated without an interval. The underlying runs would support one, and it should be reported.

\textbf{Unmeasured quantities.} We report engine file size on disk, not peak GPU memory during inference, so our claims about the 8\,GB budget rest on system-level rather than per-engine memory accounting. We did not measure Diffusion Policy's inference latency, did not record a failure-mode taxonomy for its 10 trials, and did not build an FP32 TensorRT engine, which leaves the precision effect and the graph-compilation effect entangled in our $6.4\times$ figure.

\textbf{Task-limited, not budget-limited, policy comparison.} We trained Diffusion Policy for 200k steps, LeRobot's full reference budget for that architecture and twice the budget we gave ACT, and it still did not converge to a usable policy. The failure is therefore not an artifact of an unfairly short run. It does not follow that we can rank the two architectures on task success: that reference budget was tuned on PushT, a simulated 2D planar pushing benchmark with a single end-effector, a single observation view, a rigid target object, and no grasping, and our task is harder along every one of these axes. A budget sufficient for an easier task is not guaranteed to be sufficient for ours, and a converged Diffusion Policy may well outperform the ACT policy we deploy, particularly on a deformable object where its multimodal action modeling is theoretically advantageous. We also did not sweep diffusion-specific hyperparameters (denoising steps, action normalization mode, prediction horizon).

\textbf{Scope.} A single task, a single object, a single embodiment (SO-101), and a single edge device (Orin Nano Super 8\,GB). Our forgiving deformable target is the specific reason our fidelity-versus-success divergence should not be generalized. We evaluate one fixed three-camera configuration and do not characterize the contribution of individual views. Our quantization results are specific to ACT and would not transfer to Diffusion Policy, whose iterative denoising would apply quantization error at every step rather than once per chunk.

\subsection{Future work}

The most direct extensions follow from our own gaps: train Diffusion Policy beyond its 200k-step reference budget, and sweep its denoising steps, action normalization mode, and prediction horizon, to establish what budget our task actually requires and to enable a convergence-limited architectural comparison; measure per-engine runtime memory; build an FP32 TensorRT engine to separate compilation gains from precision gains; and report dispersion across the five sensing-benchmark runs per condition.

Three larger directions follow from the findings themselves. First, \textbf{deploy ACT with temporal ensembling under FP16}, which our latency data says is now affordable and which Section~\ref{sec:conditional} identifies as the configuration where quantization changes from optional to necessary. Second, \textbf{pursue INT8 quantization of ACT's transformer layers} through explicit Q/DQ insertion (ModelOpt or equivalent), since our 0/145 result shows general-purpose calibration will not reach them; this would also test whether the coordination-augmented variants of Section~2.3, which add transformer capacity specifically, are quantizable at all. Third, \textbf{lift the sensor bottleneck}: our control loop is capped at 10\,Hz by three USB cameras sharing bus bandwidth, and CSI cameras or a higher-bandwidth capture path would raise the ceiling that everything else in this system is currently sitting under.

\section*{Acknowledgment}

This research was conducted at the Georgia Tech Research Institute (GTRI), within the Aerospace, Transportation and Advanced Systems Laboratory (ATAS). The authors thank Nathan Damen, Research Engineer at GTRI ATAS, for his mentorship and technical guidance throughout the design, implementation, and evaluation of this system. The authors also thank GTRI ATAS for providing the robotics hardware, laboratory space, and compute resources that made this work possible.

\balance
\bibliographystyle{IEEEtran}
\bibliography{references}

\end{document}